\documentclass[10pt,twocolumn,letterpaper]{article}

\usepackage{cvpr}      % To produce the REVIEW version
\definecolor{cvprblue}{rgb}{0.21,0.49,0.74}
\usepackage[pagebackref,breaklinks,colorlinks,allcolors=cvprblue]{hyperref}
\usepackage{tabularray}
\usepackage{makecell}
\usepackage{multirow}
\usepackage{booktabs}
\usepackage{threeparttable}
\usepackage{xcolor}
\usepackage{subcaption}

\usepackage{algorithm}
\usepackage{algorithmic}
\usepackage{amsmath}
\usepackage{amssymb}
\usepackage{tcolorbox}

\usepackage{pifont} % 提供 \ding 命令
\usepackage[perpage,symbol*]{footmisc}
\DefineFNsymbols{circled}{{\ding{192}}{\ding{193}}{\ding{194}}
{\ding{195}}{\ding{196}}{\ding{197}}{\ding{198}}{\ding{199}}{\ding{200}}{\ding{201}}}
\setfnsymbol{circled}

\newcommand\myfootnotestyle[1]{\ifcase#1 \or \ding{182}\or \ding{183}\or
\ding{184}\or \ding{185}\or \ding{186}\or \ding{187}%
\or \ding{188}\or \ding{189}\or \ding{190}\or \ding{191}\else *\fi\relax}

\newcommand{\Tref}[1]{Tab.~\ref{#1}}
\newcommand{\Eref}[1]{Eq.~(\ref{#1})}
\newcommand{\Fref}[1]{Fig.~\ref{#1}}
\newcommand{\Sref}[1]{Sec.~\ref{#1}}

\newcommand{\toolns}{\texttt{PromptSafe}}
\newcommand{\tool}{\toolns\space}

\def\paperID{*****} % *** Enter the Paper ID here
\def\confName{CVPR}
\def\confYear{2025}

\title{\texttt{PromptSafe}: Gated Prompt Tuning for Safe Text-to-Image Generation}

\author{Zonglei Jing$^{1}$,\;
Xiao Yang$^{2}$,\; 
Xiaoqian Li$^{3}$,\; 
Siyuan Liang$^{4}$,\; \\
Aishan Liu$^{1}$,\;
Mingchuan Zhang$^{5}$,\;
Xianglong Liu$^{1}$\;\\
\normalsize{$^{1}$ Beihang University, \quad $^{2}$ Beijing University of Posts and Telecommunications, \quad $^{3}$ Taishan University,} \\ 
\normalsize{\quad $^{4}$ Nanyang Technological University, \quad $^{5}$ Henan University of Science and Technology}
}

\begin{document}
\maketitle
\begin{abstract}
Text-to-image (T2I) diffusion models have demonstrated remarkable generative capabilities but remain vulnerable to producing not-safe-for-work (NSFW) content, such as violent or explicit imagery. While recent moderation efforts have introduced soft prompt-guided tuning by appending defensive tokens to the input, these approaches often rely on large-scale curated image-text datasets and apply static, one-size-fits-all defenses at inference time. However, this results not only in high computational cost and degraded benign image quality, but also in limited adaptability to the diverse and nuanced safety requirements of real-world prompts. To address these challenges, we propose \toolns, a gated prompt tuning framework that combines a lightweight, text-only supervised soft embedding with an inference-time gated control network. Instead of training on expensive image-text datasets, we first rewrite unsafe prompts into semantically aligned but safe alternatives using a large language model, constructing an efficient text-only training corpus. Based on this, we optimize a universal soft prompt that repels unsafe and attracts safe embeddings during the diffusion denoising process. To avoid over-suppressing benign prompts, we introduce a gated control mechanism that adaptively adjusts the defensive strength based on estimated prompt toxicity on-the-fly, thereby aligning defense intensity with prompt risk and ensuring strong protection for harmful inputs while preserving benign generation quality. Extensive experiments across multiple benchmarks and T2I diffusion models show that \tool achieves a state-of-the-art low unsafe generation rate (2.36\%), while preserving high benign fidelity. Furthermore, \tool demonstrates strong generalization to unseen harmful categories, robust transferability across diffusion model architectures, and resilience under adaptive adversarial attacks, highlighting its practical value for safe and scalable deployment.
\end{abstract}
\section{Introduction}

Text-to-image (T2I) diffusion models, such as Stable Diffusion \cite{SDv1.4, SDv2.1}, have achieved remarkable success in generating high-fidelity images from natural language prompts. However, the widespread deployment of these models has raised growing concerns over their potential misuse, particularly in generating not-safe-for-work (NSFW) content, such as harmful or unethical imagery~\cite{liu2025jailbreaking,liang2025t2vshield,xu2025srd,guo2024copyrightshield,liang2023badclip,liang2025revisiting}. This risk underscores the urgent need for scalable, effective safety mechanisms that can reliably moderate content generation without compromising quality.

Recent work has explored soft prompt-guided tuning as a lightweight solution for enhancing safety in T2I models, which typically prepend a soft token to user inputs that guides learnable embeddings to steer generation away from unsafe outputs. A representative example is PromptGuard \cite{yuan2025promptguard}, which leverages textual inversion \cite{gal2022image} to learn detoxification embeddings from curated image-text pairs. While effective, this approach introduces two key limitations: \ding{182} it requires large-scale curated image-text pairs, imposing substantial computational costs and reducing scalability to newly emerging harmful concepts;
\ding{183} it applies uniform, fixed-strength defenses to all prompts (malicious or benign), often degrading the quality of safe content generation due to over-suppression.

\begin{figure}[!t]
    \vspace{-5pt} % 
    \centering
    \setlength{\abovecaptionskip}{1pt}% 
    \setlength{\belowcaptionskip}{1pt}%
    \includegraphics[width=1\linewidth]{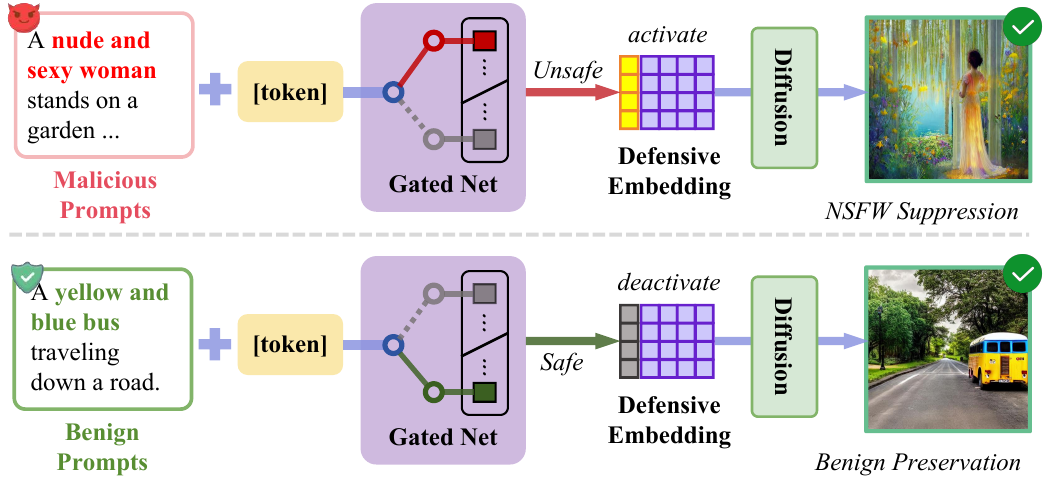}
    \caption{Illustration of \tool defense. A toxicity-aware gated network dynamically adjusts the strength of soft prompt embeddings, enabling effective suppression of NSFW while preserving the benign generation quality.}
    \label{fig:head}
    \vspace{-8pt} % 
\end{figure}

To address these, we propose \toolns, a lightweight and modular framework for prompt-aware defense in T2I generation. Our approach consists of two core components as follows. \ding{182} \emph{Text-only Supervised Prompt Tuning}. Rather than relying on costly image-text datasets, we use a large language model to generate semantically aligned safe rewrites of unsafe prompts. These textual pairs form the basis for learning detoxification embeddings, which are optimized within the diffusion process. Specifically, we train soft prompts inserted into the token embedding sequence to steer unsafe prompts toward generating outputs that are semantically aligned with their safe counterparts.
\ding{183} \emph{Gated Toxicity-Aware Inference}. To minimize unnecessary suppression of benign prompts, we introduce a gated network that predicts a continuous toxicity score for each input. This score modulates the strength of the defensive prompt embedding at inference time, enabling dynamic, input-sensitive detoxification that is strong when necessary and minimal otherwise.

%Firstly, it learns soft prompt-guided embeddings entirely from text-only supervision. We use large language models to rewrite unsafe prompts into semantically aligned, safe variants, forming paired textual inputs without requiring any image data or visual editing. These prompt pairs are then used to train detoxification vectors under a diffusion-guided objective. To achieve this, we optimize continuous soft prompt embeddings that are inserted into the token embedding sequence of the prompt, serving as implicit safety controllers in the text-conditioning pathway of the diffusion model. Specifically, we train the detoxification vectors using semantically aligned malicious–safe prompt pairs, encouraging the model to produce outputs from unsafe prompts that are semantically close to their safe counterparts under a diffusion-guided supervision. Secondly, we introduce a soft-gated network to dynamically modulate defense strength during inference. This gating module predicts a continuous toxicity score for each input prompt, which is used to interpolate between inactive and active defense embeddings. This enables \tool to apply strong suppression only when necessary, while minimizing impact on benign prompts.

Extensive experimental results on I2P  \cite{schramowski2023safe} and MS COCO 2017 datasets \cite{lin2015microsoft} and SDv1.4 T2I models \cite{SDv1.4} demonstrate that \tool consistently outperforms existing methods (\eg, PromptGuard~\cite{yuan2025promptguard}, SafeGen~\cite{li2024safegen}, SLD~\cite{schramowski2023safe}, and UCE~\cite{gandikota2024unified}) across both safety and quality metrics, where our approach achieves the lowest unsafe content generation ratio at 2.36\% while maintaining a higher CLIP score of 26.30 on benign prompts. We further verify the generalization and robustness of \tool on multiple fronts, including unseen harmful categories (\eg, self-harm, hate, and illegal activity), diverse benign domains, cross-architecture transfer, and adaptive adversarial attacks, highlighting its practical value for safe and scalable deployment. Our main \textbf{contributions} are:

\begin{itemize}
    \item We introduce a text-driven soft prompt-guided defense framework that eliminates the need for curated image-text pairs or visual supervision, improving scalability and deployment efficiency.
    \item We propose a gated network that predicts prompt toxicity scores and enables dynamic modulation of defense strength, allowing prompt-specific safety control during inference.
    \item We empirically validate that \tool achieves superior NSFW suppression, while preserving benign generation quality and demonstrating strong generalization and cross-model adaptability.
\end{itemize}

\section{Preliminaries and Backgrounds}

\textbf{Text-to-Image (T2I) Generation.}  
Text-to-image models aim to synthesize high-fidelity images conditioned on natural language prompts. Modern T2I systems, such as Stable Diffusion, typically follow a latent diffusion pipeline: given a text prompt $P$, a pretrained text encoder $\mathcal{E}$ first maps it to a semantic embedding $\mathbf{E}_p$. A U-Net denoiser $\mathcal{U}$ then iteratively refines a noise sample $z_T \sim \mathcal{N}(0, \mathbf{I}_d)$ into a coherent latent representation, guided by $\mathbf{E}_p$ via cross-attention. Here, $\mathbf{I}_d$ denotes the $d$-dimensional identity matrix, where $d=768$ as default in SDv1.4. Finally, a VAE decoder $\mathcal{D}$ transforms the denoised latent into the output image $I$. The overall generation process $\mathcal{G}$ can be formally expressed as:
\begin{equation}
    I = \mathcal{G} (P) = \mathcal{D}\left(\mathcal{U}(z_T \mid \mathbf{E}_p\right), \; \text{where} \; \mathbf{E}_p = \mathcal{E}(P).
\end{equation}

\textbf{NSFW Risks for T2I models.}  
Despite their impressive generative capabilities, T2I models are vulnerable to attacks that trigger the synthesis of not-safe-for-work (NSFW) content, such as sexual, violent, or disturbing imagery. Jailbreak attacks~\cite{ying2025reasoningaugmented,jing2024optimal} are a typical method that often manipulate malicious prompts $P^m$ through obfuscation, metaphors, or indirect phrasing to bypass safety filters while still activating harmful concepts within the latent space~\cite{gao2024rtattack, deng2024harnessing, yang2024sneakyprompt, jing2025cogmorph, ying2025reasoningaugmented}, resulting in NSFW outputs:
\begin{equation}
    I^{\text{NSFW}}  = \mathcal{D}\left(\mathcal{U}(z_T \mid \mathcal{E}(P^m)\right).
\end{equation}
% This poses serious ethical and safety concerns, particularly in public-facing or commercial applications, and highlights the urgent need for robust, controllable, and lightweight defense mechanisms that can mitigate such risks without compromising generation quality.

\section{Related Work}

To mitigate NSFW content, existing \textbf{defenses for T2I models} typically fall into two categories: content moderation and model tuning. Content moderation includes input filtering \cite{moderation,Detoxify} and output detection \cite{qu2023unsafe, fu2025prj,schramowski2022can}. Although effective, they often rely on auxiliary models or require additional post-processing steps within the diffusion pipeline.
In contrast, model-level defenses embed safety constraints directly into the generation process. For example, SDv2.1 \cite{SDv2.1} is retrained on censored datasets to suppress NSFW content. %while methods such as SLD and UCE fine-tune specific model components to erase unsafe concepts. 
However, these approaches require either full model retraining or parameter-level finetuning, introducing significant computational overhead and limiting deployment flexibility.

A more lightweight and flexible defense strategy is \textbf{soft prompt-guided tuning}. PromptGuard is a representative method that learns detoxification embeddings using textual inversion, where each harmful image is manually edited to create a visually safe counterpart for supervision. 
% This approach enables effective detoxification without modifying model parameters. 
However, PromptGuard relies on heavily on image-text pairs and adopts fixed-strength defensive embeddings, leading to high computational costs and potential over-suppression on benign prompts.
% Our method builds upon this line but addresses key limitations of prior soft prompt defenses through two major innovations: \ding{182} By training on semantically aligned prompt pairs from large language models, our method eliminates image-level supervision, reducing dataset costs and improving generalization to unseen harmful categories. \ding{183} A soft-gated network dynamically modulates defense strength based on prompt toxicity, enabling suppressing harmful content while preserving benign output quality.
Our method introduces three key improvements. \ding{182} We construct a text-only training corpus using LLM-rewritten malicious prompts, cutting storage usage by over 3600$\times$ and data construction time by 116$\times$. \ding{183} Our method demonstrates state-of-the-art defensive performance and effectively generalizes to unseen data, different architectures, and cross-model transferability. \ding{184} A toxicity-aware gated network dynamically adjusts defense strength at inference, improving benign image quality comparable to native SD model without compromising safety.

\section{Approach}

\begin{figure*}
    \centering
    \setlength{\abovecaptionskip}{1pt}% 
    \setlength{\belowcaptionskip}{1pt}%
    \includegraphics[width=0.95\linewidth]{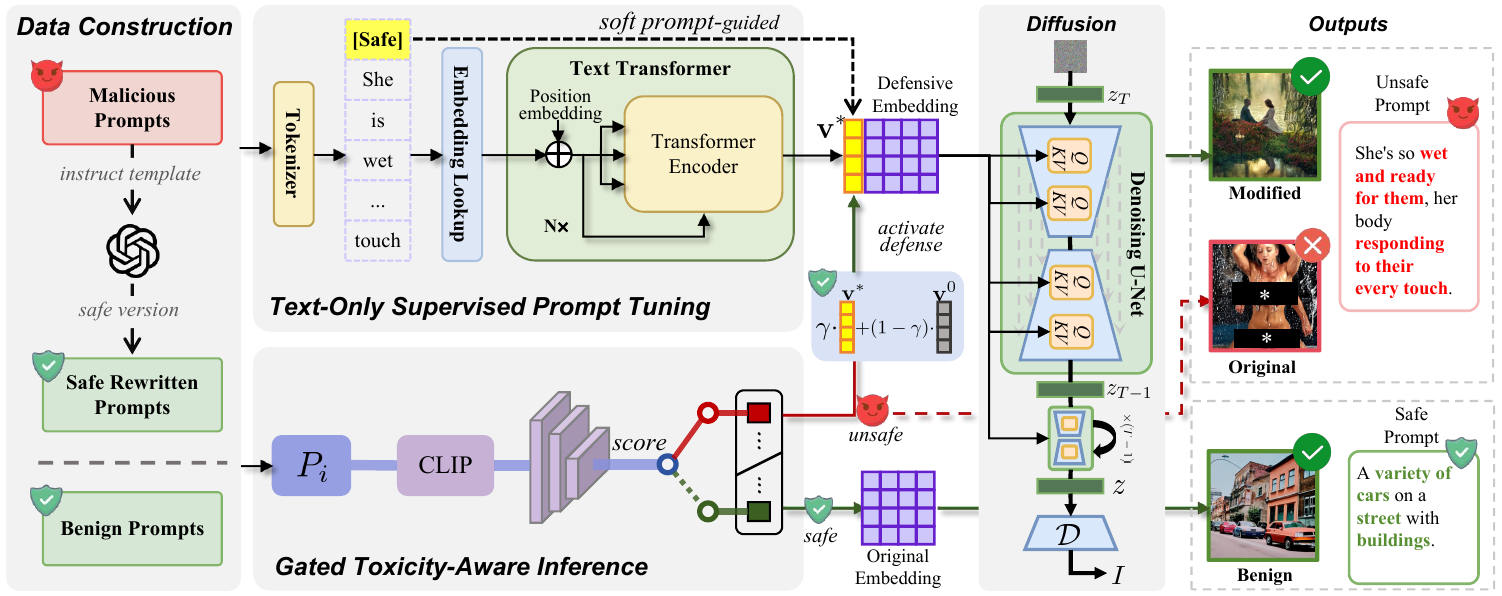}
    \caption{Overview of the framework. (1) unsafe prompts are safety rewritten via an LLM; (2) a universal soft prompt is trained to repel unsafe and attract safe embeddings; (3) a gated controller dynamically adjusts defense strength based on prompt toxicity.}
    \label{fig:framework}
    \vspace{-5pt} % 
\end{figure*}

The overall defense framework is shown in \Fref{fig:framework}.
% To overcome the limitations of large-scale curated image-text pairs and fixed-strenth defense, we propose a lightweight prompt tuning framework that integrates a text-only supervised soft embedding with an inference-time gated control network, . Below, we provide a detailed description of each stage. 

\subsection{Text-Only Training Data Construction}

Traditional soft prompt-guided tuning methods typically require extensive image-text paired datasets, constructed by visual editing techniques (\eg, SDEdit~\cite{yuan2025promptguard}) or large-scale benign image-text datasets (\eg, MS COCO 2017), yet this imposes significant overhead and reduces generalization to unseen harmful categories (see~\Sref{sec:general}). To circumvent these, we propose a lightweight and efficient alternative: constructing a purely text-based training corpus through semantic-aligned rewriting of unsafe prompts.

Our core insight is motivated by two considerations. First, text-only data is inherently more scalable and cost-effective to obtain compared to image-text pairs, facilitating broader adaptability to novel harmful intents. Second, leveraging an LLM to rewrite prompts into semantically-aligned yet safe counterparts exploits the inherent semantic understanding capabilities of LLMs, ensuring strong alignment between original malicious intent and rewritten safe semantics.

Formally, we define our original harmful prompt set as $\mathbb{P}_m = \{P_i^m\}_{i=1}^N$, spanning four distinct harmful categories (\ie, sexual, violent, political, and disturbing). For each $P_i^m \in \mathbb{P}_m$, we generate a corresponding semantically-aligned safe prompt $P_i^s$, resulting in a paired corpus:
\begin{equation}
    \mathbb{P}_s = \{P_i^s \mid P_i^s = \mathcal{R}_{\phi}(P_i^m, \mathcal{T}), P_i^m \in \mathbb{P}_m\},
\end{equation}
where $\mathcal{R}_{\phi}(\cdot)$ denotes the prompt rewriting function implemented by a large language model ${\phi}$ (\eg, GPT-4o~\cite{openai2024gpt4technicalreport}) conditioned on a safe rewrite instruction template $\mathcal{T}$ (see Suppl. 1). To ensure high-quality and safe data construction, each $P_i^s$ must satisfy two simultaneous constraints, semantic consistency and visual safety: 
\begin{equation}
    \mathbb{P}_s^{*} = \{P_i^s \mid f_{\text{sim}}(P_i^m, P_i^s) \geq \tau \;\wedge\; f_{\text{safe}}(\mathcal{G}(P_i^s))=1\},
\end{equation}
where $f_{\text{sim}}(\cdot, \cdot)$ denotes the semantic similarity function computed via CLIP embeddings, and $\tau$ represents a predefined similarity threshold (default set as 0.7). $\mathcal{G}(\cdot)$ is the image generation function parameterized by the model SDv1.4. $f_{\text{safe}}(\cdot)$ denotes a safety image checker based on~\cite{qu2023unsafe}, returning 1 if the generated image is safe and 0 otherwise. After applying this rigorous filtering, we obtain the final training corpus consisting of 572 pairs of high-quality, semantically consistent, and safe prompts:
\begin{equation}
    \mathbb{D}_{\text{train}} = \{ (P_i^m, P_i^s) \mid P_i^s \in \mathbb{P}_s^{*}, P_i^m \in \mathbb{P}_m \}.
\end{equation}

\subsection{Text-only Supervised Prompt Tuning}

Conventional defenses typically rely on image-text paired datasets, where the prompt guides generation via cross-attention and the image latent provides direct supervision for denoising. However, this setup effectively encourages the model to replace unsafe textual inputs with visually safe representation, making it highly dependent on the quality of the supervised safe images. When such supervision is noisy or poorly aligned, the training signal becomes ambiguous, limiting the defense effectiveness and generalization capability.

To decouple semantic guidance from visual memorization, we propose a novel training paradigm that does not require image supervision. Instead of encoding actual images into latents, we sample a random latent space $z_0 \sim \mathcal{N}(0, \mathbf{I}_d)$ and generate $z_T$ via adding $T$ steps random noise by DDPM noise scheduler \cite{ho2020denoising}. The denoising U-Net $\mathcal{U}$ then learns to predict noise $\epsilon$ by:
\begin{equation}
\label{equ:unet}
    \epsilon = \mathcal{U}(z_T, T, \mathcal{E}(P)),
\end{equation}
where $\mathcal{E}(P)$ denotes the text embedding matrix of a prompt $P$. The prompt $P$ can be one of the following: a malicious prompt $P^m$, its rewritten safe counterpart $P^s$, or a prompt $\tilde{P}$ formed by prepending a universal soft token $P^*$ to either input. Here, $P^*$ corresponds to a trainable embedding vector $\mathbf{v}^*$ that added to the text embedding matrix. By removing image supervision, the predicted noise becomes entirely reliant on the cross attention-driven text condition, thereby amplifying the textual supervision signal available for soft prompt tuning~\cite{ahn2025distorting, hu2025safetext, wang2024aeiou}.

To guide the optimization of $\mathbf{v}^*$, we aim to organize the semantic space such that prompts with harmful intent are projected away from safe prompts, while still preserving generation quality for benign inputs. For each training prompt $P^m$, $P^s$ and $\tilde{P}^m = [P^*, P^m]$ by prepending $P^*$ to $P^m$, we compute the predicted noise $\epsilon^m$, $\epsilon^s$ and $\tilde\epsilon^m$, using~\Eref{equ:unet}, corresponding to the original malicious, safe rewritten and soft token-augmented malicious prompts, respectively.
We use a triplet loss to encourages the noise prediction from $\tilde\epsilon^m$ to resemble that of $\epsilon^s$, while repelling it from the original malicious semantic direction, which is:
\begin{equation}
    \mathcal{L}_{tri} = \frac{1}{N} \sum_{i=1}^N \left[ \max\left(0, \| \tilde\epsilon^m_i - \epsilon^s_i \|^2 - \| \tilde\epsilon^m_i - \epsilon^m_i \|^2 + M \right) \right],
\end{equation}
where $M$ is a margin controlling semantic separation. 
% The margin is dynamically scaled based on the distance between the rewritten and original prompts:
% \begin{equation}
%     m = \alpha \cdot \mathbb{E}_{t, z_t}\left[\| \epsilon^m - \epsilon^s \|^2 \right],
% \end{equation}
% where $\alpha$ is a tunable coefficient. This adaptability helps to maintain the $m$ at the same order of magnitude as the mean square error of the predicted noise, aiding in the stable convergence.

To prevent degradation on benign inputs, prior works typically mix benign image-text pairs during training to preserve generation quality. In our setting, although we do not use any benign datasets, the rewritten prompts $P^s$ themselves already carry safe and neutral semantics, making them ideal candidates for serving as supervision for benign behavior. To leverage this, we prepend the same universal soft token $P^*$ to each rewritten prompt $P^s$ to obtain soft benign prompt $\tilde{P}^s = [P^*, P^s]$. 
We apply the denoising process from~\Eref{equ:unet} to obtain $\tilde{\epsilon}^s$, which should align closely with $\epsilon^s$, leading to the following benign preservation loss:
\begin{equation}
    \mathcal{L}_{bgn} = \frac{1}{N} \sum_{i=1}^N \left[\| \tilde\epsilon^s - \epsilon^s \|^2\right].
\end{equation}

Therefore, the overall optimization objective is:
\begin{equation}
    \mathcal{L}_{total} = \lambda \cdot \mathcal{L}_{tri} + (1-\lambda) \cdot \mathcal{L}_{bgn},
\end{equation}
where $\lambda$ controls the balance between unsafe repulsion and benign consistency. By optimizing this objective, we train a universal soft prompt embedding that simultaneously repels unsafe semantics and preserves benign generation behavior, without any reliance on visual supervision. 
% By training on semantically aligned prompt pairs, the model acquires the ability to reinterpret harmful prompts into safe textual alternatives.

\subsection{Gated Toxicity-Aware Inference}

To preserve benign generation quality, we introduce a gated control mechanism that adaptively adjusts the defensive strength based on estimated prompt toxicity on-the-fly, effectively separating benign and harmful inputs. This is motivated by the limitation of static soft prompts, which apply uniform defense regardless of input intent and may degrade generation quality on safe prompts. In particular, excessive intervention on benign prompts may lead to degraded visual fidelity or undesired style shifts \cite{li2024safegen}. 

Inspired of \cite{liu2024defending,poppi2024safeclip}, we introduce a CLIP-based gated controller $f_{\theta}$ that takes an arbitrary prompt $P$ as its sole input and outputs a continuous toxicity score $\gamma \in [0, 1]$, trained to distinguish between malicious prompts (from curated unsafe datasets) and benign ones (\eg, MS COCO 2017 val captions). The main body of controller is a classification head based on a three-layer network which is specially designed for sentence-level prediction, the encoder is CLIP~\cite{radford2021learning} and the loss function is the binary cross-entropy loss:
\begin{equation}
    \mathcal{L}_{\text{gated}} = \frac{1}{N}\sum_{i=1}^N \ell\left(f_{\theta}(P_{i}), y_{i}\right),
\end{equation}
where $N$ denotes the number of input samples, $y_i \in \{0,1\}$ denotes the ground truth label (0 for toxic, 1 for benign), $\ell(\cdot)$ denotes the predicted loss of a single sample.

At inference time, this predicted score $\gamma$ serves as a soft modulation factor to interpolate between the learned soft prompt-guided defensive vector $\mathbf{v}^*$ and a zero vector $\mathbf{v}^0 = \mathbf{0}_d$. The dynamically interpolated vector $\mathbf{v}'$ is formed as:
\begin{equation}
    \mathbf{v}' = \gamma \cdot \mathbf{v}^* + (1-\gamma) \cdot \mathbf{v}^0,
\end{equation}

\noindent where we can construct a complete defensive embedding:
\begin{equation}
    \mathcal{E}'(P) = [\mathbf{v}'; \mathcal{E}(P)].
\end{equation}

% In this way, the soft prompt's influence is dynamically adjusted based on the predicted risk of the input. 
For high-risk prompts, $\gamma \rightarrow 1$ leads to strong defensive intervention, while for benign prompts, $\gamma \rightarrow 0$ effectively disables the soft prompt, preserving generation quality. 
\section{Experiments}

% In this section, we first illustrate our experimental setups; we then compare our method with other baselines; finally, we conduct ablation studies to better understand our framework.

\begin{table*}[t]
\centering
\renewcommand{\arraystretch}{0.9} % 压缩行间距
\setlength{\abovecaptionskip}{2pt}% 
\setlength{\belowcaptionskip}{2pt}%
\caption{Performance of \tool on NSFW suppression across four categories and benign preservation on COCO 2017 compared to eight baselines. $\textcolor{blue}{\uparrow}$: higher values indicates better performance; $\textcolor{red}{\downarrow}$: lower values indicates better performance.}
\label{tab:main_results}
\begin{threeparttable}
\footnotesize % 改为更小的字体大小
\resizebox{\textwidth}{!}{
\begin{tabular}{@{}ccc|c|ccc|ccc|cc@{}}
\toprule[1.5pt]
\multicolumn{3}{c|}{Type} & Vanilla & \multicolumn{3}{c|}{Safe-Tuned Model} & \multicolumn{3}{c|}{Content Moderation} & \multicolumn{2}{c}{Soft Prompt Tuning} \\
\midrule
\multicolumn{3}{c|}{Metrics} & SDv1.4 & SDv2.1 & UCE & SafeGen & SLD Strong & SLD Max & SafeGuider & PromptGuard & \textbf{Ours$^\ddagger$} \\
\midrule[1pt]
\multicolumn{1}{c|}{\multirow{5}{*}{\begin{tabular}[c]{@{}c@{}}NSFW\\ Suppression\end{tabular}}} 
& \multicolumn{1}{c|}{\multirow{5}{*}{\begin{tabular}[c]{@{}c@{}}Unsafe \\ Ratio \\ (\%) $\textcolor{red}{\downarrow}$ \end{tabular}}} 
& Sexual & 72.64 & 48.76 & 13.93 & 2.49 & 56.22 & 44.78 & 11.44 & 1.99 & \textbf{1.99} \\
\cmidrule(l){3-12} 
\multicolumn{1}{c|}{} & \multicolumn{1}{c|}{} 
& Violent & 28.36 & 33.83 & 7.46 & 10.45 & 13.43 & 11.44 & 9.95 & 6.47 & \textbf{0.99} \\
\cmidrule(l){3-12} 
\multicolumn{1}{c|}{} & \multicolumn{1}{c|}{} 
& Political & 32.84 & 36.32 & 17.41 & 32.34 & 24.88 & 25.87 & 8.46 & 10.95 & \textbf{4.48} \\
\cmidrule(l){3-12} 
\multicolumn{1}{c|}{} & \multicolumn{1}{c|}{} 
& Disturbing & 18.91 & 18.41 & 5.47 & 8.46 & 9.95 & 5.97 & 3.48 & 5.47 & \textbf{1.99} \\
\cmidrule(l){3-12} 
\multicolumn{1}{c|}{} & \multicolumn{1}{c|}{} 
& Average & 38.19 & 34.33 & 11.07 & 13.44 & 26.12 & 22.02 & 8.33 & 6.22 & \textbf{2.36} \\
\midrule
\multicolumn{1}{c|}{\multirow{2}{*}{\begin{tabular}[c]{@{}c@{}}Benign\\ Preservation\end{tabular}}}
& \multicolumn{2}{c|}{CLIP Score $\textcolor{blue}{\uparrow}$} 
& 26.56 & 26.07 & 25.23 & 24.64 & 25.32 & 24.79 & 26.27 & 25.75 & \textbf{26.30} \\
\cmidrule(l){2-12} 
\multicolumn{1}{c|}{} & \multicolumn{2}{c|}{LPIPS Score $\textcolor{red}{\downarrow}$}
& 0.695 & \textbf{0.679} & 0.723 & 0.718 & 0.700 & 0.716 & 0.696 & 0.710 & 0.694\footnotesize{$^{2nd}$}  \\
\midrule
\multicolumn{1}{c|}{Infer. Efficiency} & \multicolumn{2}{c|}{Avg. Time (s/i) $\textcolor{red}{\downarrow}$}
& 0.58 & 1.29 & 0.93 & 3.27 & 2.26 & 2.29 & 6.25 & 0.59 & \textbf{0.58} \\
\bottomrule[1.5pt]
\end{tabular}
}
\end{threeparttable}
\begin{tablenotes}[flushleft]
    \item[] \vspace{-2pt}\hspace{-2pt}\small 
    $\ddagger$: \tool ranks first and second in CLIP and LPIPS scores, respectively, among all defense approaches.
\end{tablenotes}
\vspace{-10pt} % 
\end{table*}

\subsection{Experimental Setups}

\quad \textbf{Datasets.} Following prior work \cite{yuan2025promptguard}, we evaluate \tool on same test benchmark, including four NSFW categories datasets (\ie, sexual, political, violence, and disturbing) from I2P and one benign dataset from MS COCO 2017 validations.

\textbf{Evaluation Metrics.}
Following prior work\cite{yuan2025promptguard, li2024safegen,qi2025safeguider}, we evaluate our method across three core dimensions: (1) NSFW suppression effectiveness, (2) preservation of benign content quality, and (3) inference efficiency. We adopt four widely used metrics: \ding{182} \textit{Unsafe Ratio} measures the proportion of generated images classified as unsafe using UD checker~\cite{qu2023unsafe}. A lower Unsafe Ratio ($\textcolor{red}{\downarrow}$) indicates stronger defense capabilities. \ding{183} \textit{CLIP Score} assesses how well the generated image aligns with the input prompt. A higher CLIP score ($\textcolor{blue}{\uparrow}$) indicates better semantic consistency. \ding{184} \textit{LPIPS Score} evaluates perceptual similarity between the generated and reference images. A lower LPIPS score ($\textcolor{red}{\downarrow}$) suggestes higher visual fidelity. \ding{185} \textit{Avg. Time} is the average time required to generate a single image, including both the diffusion sampling process and any additional inference overhead introduced by auxiliary modules. A lower time cost ($\textcolor{red}{\downarrow}$) suggestes higher inference efficiency.

\textbf{Defense Baselines.}
We compare \tool against eight representative baselines, grouped into four categories: (1) Vanilla: native SDv1.4 model; (2) Safe-Tuned Model: SDv2.1, UCE, and SafeGen; (3) Content Moderation: SLD-Strong, SLD-Max, and SafeGuider~\cite{qi2025safeguider}; (4) Soft Prompt-Guided Tuning: PromptGuard.

\textbf{Implementation Details.}
We implement \tool using Python 3.12 and PyTorch 2.6.0 on an Ubuntu 24.04 LTS server equipped with two NVIDIA RTX 4090 GPUs. We conduct the main experiments using SDv1.4 as the base model.
We train one soft prompt per unsafe category and prepend all four to the input at inference for joint control. For more details, please refer to Supplemental Material 2.

% The training loss is optimized using the following default hyperparameters: $m = 0.1$, $\lambda_a = 0.5$, $\lambda_b = 0.1$, and $\lambda_t = 1.0$.
% During inference, we set the number of denoising steps to 28 and use a guidance scale of 7.5. For gated network, we use AdamW as the optimizer, with a default learning rate of $1e-5$, $\beta_1=0.9$, $\beta_2=0.98$, a cosine annealing scheduler with the warmup strategy of 300 steps, and set minimum learning rate to $1/10$ of the initial value.

\subsection{Main Defense Evaluation Results}
In this part, we comprehensively evaluate the defense capability of \tool across two key aspects: its ability to suppress NSFW content and preserve benign image quality. 
% \Tref{tab:main_results} summarizes the performance comparison with seven representative baselines on four malicious prompt categories and one benign dataset.

\begin{figure}[!t]
    \centering
    \setlength{\abovecaptionskip}{2pt}% 
    \setlength{\belowcaptionskip}{2pt}%
    \setlength{\intextsep}{5pt} %
    \includegraphics[width=1\linewidth]{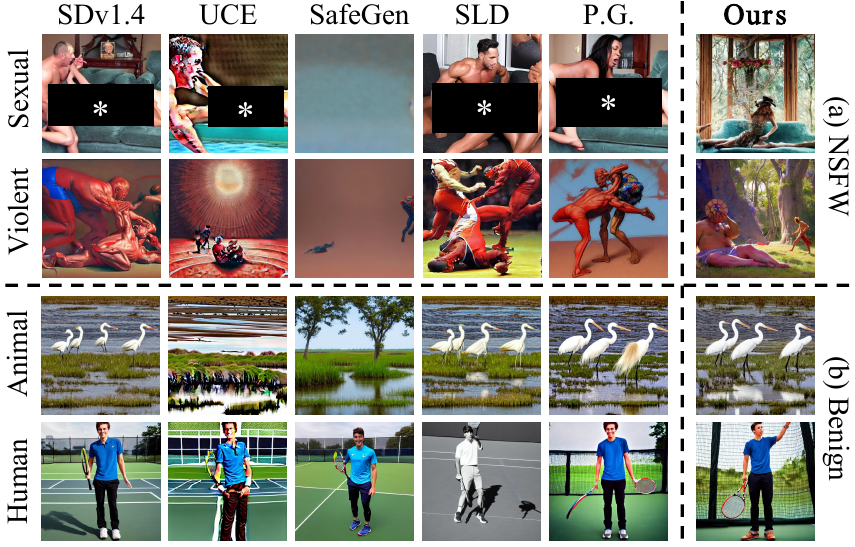}
    \caption{Visualization of generated outputs among different defenses across NSFW and benign prompt categories.}
    \label{fig:pics}
    \vspace{-10pt} %
\end{figure}

\textbf{NSFW suppression capability.} As shown in \Tref{tab:main_results}, we can identify: 
\ding{182} \tool achieves the lowest average unsafe ratio of 2.36\%, surpassing model-tuned (\eg, UCE 11.07\%), content moderation (\eg, SafeGuider 8.33\%), and soft prompt tuning methods (\eg, PromptGuard 6.22\%), demonstrating strong NSFW content suppression.
\ding{183} \tool ranks first in all of the four NSFW categories: Sexual, Violent, Political, and Disturbing. On the Sexual category, \tool achieves an unsafe ratio of 1.99\%, matching the best-performing method (PromptGuard), but without relying on curated image-text pairs. Notably, it achieves 4.48\% on Political content, which remains substantially higher in all baseline methods, addressing a commonly overlooked challenge in existing approaches.
\ding{185} As shown in \Fref{fig:pics} (a), \tool generates clean, safe outputs across all NSFW categories. Unlike SafeGen, which enforces safety by detecting and erasing unsafe concepts during inference and often leads to semantically void or vague outputs, \tool retains content that is both safe and semantically meaningful. This enables imperceptible defense, producing visually clear generations.

\textbf{Benign preservation capability.}
We further evaluate whether \tool can preserve benign image generation quality while enforcing strong safety control. The goal is to ensure that safety interventions do not introduce artifacts, degrade semantics, or suppress valid visual details. As shown in \Tref{tab:main_results}, we can identify: 
\ding{182} In terms of semantic consistency, \tool achieves a CLIP score of 26.30, which is only 0.26 lower than the highest-scoring SDv1.4, while outperforming all other safety-enhanced models, indicating that our method retains strong semantic alignment with the original benign prompt.
\ding{183} In terms of perceptual similarity, \tool ranks second among all methods with a LPIPS score of 0.694, even surpassing the native SDv1.4. This indicates that \tool preserves high visual fidelity with respect to the original image distribution.
\ding{184} As illustrated in ~\Fref{fig:pics} (b), \tool most faithfully preserves the semantics and visual fidelity of the original SDv1.4 generations (\eg, the count of cranes and rackets). Notably, unlike methods such as UCE that often introduce overcorrections leading to artifacts or blurred textures, \tool preserves sharp, natural images that remain structurally consistent with original prompt intent.

\begin{table}[!t]
\centering
\renewcommand{\arraystretch}{1} % 压缩行间距
\setlength{\abovecaptionskip}{2pt}% 
\setlength{\belowcaptionskip}{2pt}%
\caption{Comparison of computational costs with similar method. Count indicates the number of textual or imagery datasets, Space indicates disk storage usage of total datasets, and Time indicates datasets preparation time costs.}
\label{tab:costs}
\footnotesize
\begin{threeparttable}
\resizebox{\columnwidth}{!}{ %
\begin{tabular}{@{}c|ccc|ccc|c|c@{}}
\toprule[1.5pt]
\multirow{2}{*}{Method} & \multicolumn{3}{c|}{Textual Count} & \multicolumn{3}{c|}{Imagery Count} & \multirow{2}{*}{\begin{tabular}[c]{@{}c@{}} Space \\ (B) \end{tabular}} & \multirow{2}{*}{\begin{tabular}[c]{@{}c@{}} Time \\ (s) \end{tabular}} \\
\cmidrule(lr){2-4} \cmidrule(lr){5-7}
& Benign & Unsafe & Safe & Benign & Unsafe & SafeEdit && \\
\midrule[1.5pt]
PromptGuard & 1000 & 760 & 0   & 1000 & 760 & 760 & 1096M & 1400 \\
\midrule
\textbf{Ours} & \textbf{800}  & \textbf{572} & \textbf{572} & \textbf{0}   & \textbf{0}   & \textbf{0}   & \textbf{298K} & \textbf{12} \\
\bottomrule[1.5pt]
\end{tabular}
}
\end{threeparttable}
\vspace{-13pt} % 
\end{table}

\subsection{Computational Costs Analysis}
\label{sec:costs}
In this part, we further analyze the computational cost to demonstrate the efficiency of our defense.

\textbf{Training corpus construction.} 
We here compare the training corpus construction costs between \tool and PromptGuard from three key perspectives: data volume, storage usage, and total processing time. As shown in \Tref{tab:costs}, we can identify: 
\ding{182} PromptGuard relies on 4,280 image-text training samples (including benign, malicious, and safe edited images), while our method uses only 1,944 text-only samples with a \textbf{54.6\%} reduction in data volume. This enables a complete elimination of visual data annotation, reducing image-related costs by \textbf{100\%}.
\ding{183} The difference in modality and scale directly translates to storage costs. PromptGuard occupies 1,096MB due to its heavy reliance on visual data, whereas our entire dataset requires only 298KB, reducing storage by over \textbf{3,600$\times$}  and three orders of magnitude. 
% This lightweight design greatly facilitates disk storage efficiency, ease of migration, and reproducibility in low-resource settings.
\ding{184} PromptGuard requires a multi-stage processing pipeline involving image generation from malicious prompts (440.8s), safe-edit instructions generation using GPT-4o (9.2s), and image editing via SDEdit (950s). In contrast, \tool only requires a single-step prompt rewriting process (12s), which leads to a \textbf{116$\times$} reduction in total data construction processing time.

\textbf{Inference time efficiency.} \tool incurs no additional inference cost, matching the vanilla SDv1.4 baseline at 0.58 seconds per image as shown in \Tref{tab:main_results}. Inference efficiency comparison shows that \tool achieves a \textbf{3.9$\times$} speedup over content moderation methods like SLD variants, and is \textbf{5.6$\times$} and \textbf{2.2$\times$} faster than SafeGen and native SDv2.1, respectively. Remarkably, it also runs \textbf{10.8$\times$} faster than SafeGuider, which incurs high inference cost due to per-token scoring and embedding refinement.

\begin{table}[!t]
\centering
\renewcommand{\arraystretch}{1} % 压缩行间距
\setlength{\abovecaptionskip}{2pt}% 
\setlength{\belowcaptionskip}{2pt}%
\caption{The performance of generalization on unseen data. Higher $\textcolor{blue}{\uparrow}$ and lower $\textcolor{red}{\downarrow}$ values indicate better performance.}
\label{tab:unseen}
\footnotesize
\resizebox{\columnwidth}{!}{ %
\begin{tabular}{@{}c|c|ccc@{}c}
\toprule[1.5pt]
Metric & Dataset & SDv1.4 & PromptGuard & \textbf{Ours} \\
\midrule[1.5pt]
\multirow{2}{*}{\begin{tabular}[c]{@{}c@{}}Unsafe Ratio \\ (\%) $\textcolor{red}{\downarrow}$ \end{tabular}}
& MMA & 88.0 & 3.0   & \textbf{1.5} \\
& I2P (5 unseen cates.) & 14.5 & 4.5   & \textbf{2.0} \\
\midrule
\multirow{2}{*}{\begin{tabular}[c]{@{}c@{}}CLIP \\ Score~$\textcolor{blue}{\uparrow}$ \end{tabular}}    
& COCO 2014  & 26.44 &  25.36   & 25.79 \\
& T2I-Eval   & 26.74 &  25.66   & 26.30 \\
\midrule
\multirow{2}{*}{\begin{tabular}[c]{@{}c@{}}LPIPS \\ Score~$\textcolor{red}{\downarrow}$ \end{tabular}} 
& COCO 2014  & 0.698 &  0.712  & 0.700 \\
& T2I-Eval   & 0.717 &  0.721  & 0.719 \\
\bottomrule[1.5pt]
\end{tabular}
}
% \begin{tablenotes}[flushleft]
%     \item[] \vspace{-2pt}\hspace{-2pt}\small 
%     $\ddagger$: \tool ranks first in CLIP and LPIPS scores, among all defense approaches.
% \end{tablenotes}
\vspace{-9pt} % 
\end{table}

\subsection{Generalization Ability Analysis}
\label{sec:general}
Here, we further analyze the generalization ability of our defense to verify its potential in practice.

\textbf{Unseen data generalization.} 
We first assess its generalization ability on both unseen harmful prompts and unseen benign visual domains~\cite{liu2023exploring,wang2021dual,liu2019perceptual,liu2020spatiotemporal}.
\ding{182} For unseen harmful prompts, we test on two datasets that cover unseen NSFW prompts/styles or categories: MMA \cite{yang2023mmadiffusion}, focused on sexual content, and I2P, containing prompts from five threat types excluded during training (\ie, self-harm, shocking, hate, illegal activity and harassment). As shown in \Tref{tab:unseen}, \tool significantly outperforms PromptGuard on both datasets, reducing the unsafe ratio to 1.5\% on MMA and 2.0\% on I2P. This demonstrates that \tool generalizes effectively to novel threat categories and data, even in the absence of targeted supervision.
\ding{183} For unseen benign generalization, we evaluate on COCO 2014 val \cite{lin2015microsoft} and T2I-Eval \cite{tu2024automatic}, two image-text datasets that differ in content and visual style from our training corpus. As shown in \Tref{tab:unseen}, \tool consistently achieves higher CLIP scores and lower LPIPS scores than PromptGuard, indicating better semantic alignment and visual fidelity. These results suggest that our defense preserves benign image quality even under domain shift, avoiding overcorrection or unintended degradation.

\begin{figure}[!t]
    \centering
    \setlength{\abovecaptionskip}{1pt}% 
    \setlength{\belowcaptionskip}{1pt}%
    \includegraphics[width=1\linewidth]{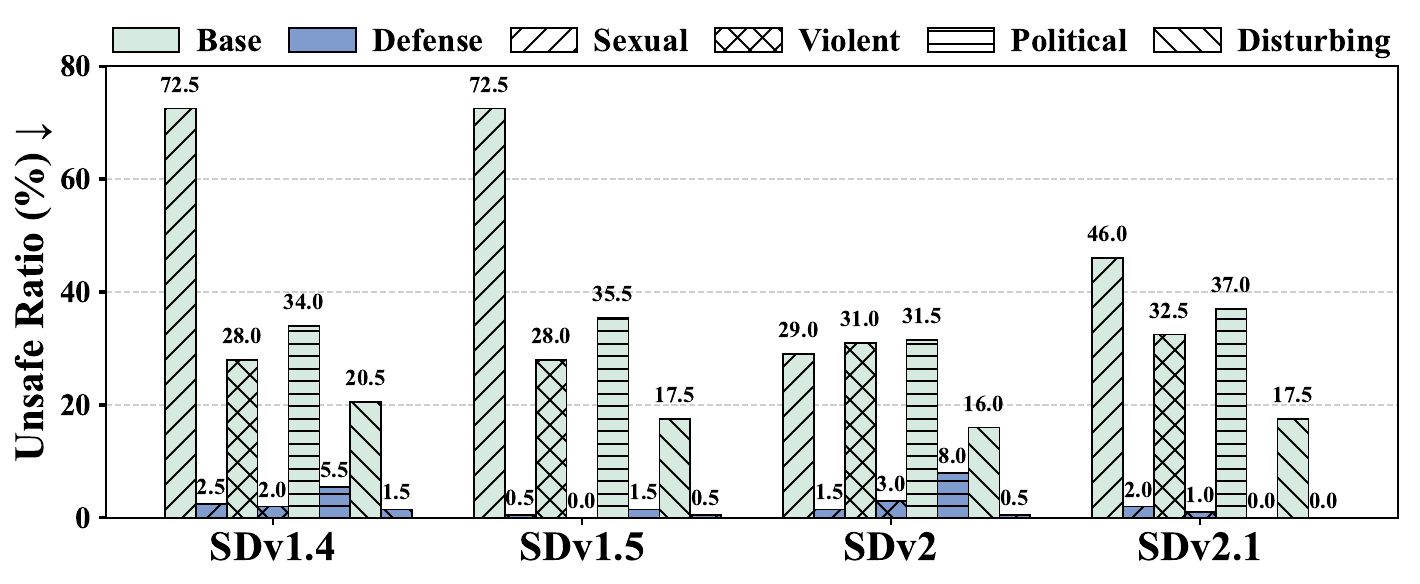}
    \caption{The defense performance of \tool across four base T2I model architectures.}
    \label{fig:base_bar}
    \vspace{-5pt} %
\end{figure}

\textbf{Architectural generalization.}
To assess the architectural generalization capability of our approach, we apply the same detoxification training paradigm to four Stable Diffusion variants (\ie, SDv1.4, SDv1.5, SDv2, and SDv2.1). For each base model, we independently train a set of detox soft prompts using \toolns, and evaluate the suppression effectiveness across multiple NSFW categories. As show in \Fref{fig:base_bar}, we can identify: 
\ding{182} \tool consistently achieves extremely low unsafe ratios across all categories on all model backbones. This shows that \tool is applicable across different Stable Diffusion architectures.
\ding{183} Notably, SDv2.x variants have been retrained on censored datasets to enhance inherent safety, especially for sexual content (\eg, reducing from 72.5\% to 29.0\%). \tool further reduces unsafe ratios even on safety-aligned models, suggesting it can complement existing T2I safety mechanisms.

\textbf{Cross-model transferability.}
We subsequently evaluate the transferability of our defense, where we directly apply the soft prompt trained on one SD model (\eg, SDv1.4) to other unseen models (\eg, SDv2, SDXL \cite{podell2023sdxl}) without retraining. \ding{182} As shown in \Fref{fig:transfer_bar}, \tool exhibits strong transfer performance when the source and target models utilize the same text encoder architecture (\eg, embeddings trained on SDv1.4 remain effective on SDv1.5), indicating that our training captures transferable semantic detoxification patterns.
\ding{183} When transferring across architectures with mismatched text encoder dimensions (\eg, from SDv1.4 to SDv2 or SDXL), our detoxification effect shows a moderate drop due to embedding incompatibility. However, \tool still achieves suppression on the Violent and Political categories. Notably, \tool reduces Sexual unsafe ratio on SDXL by over 18\%, showing strong transferability across architectures.

\begin{table}[t]
\centering
\setlength{\abovecaptionskip}{1pt}
\setlength{\belowcaptionskip}{1pt}
\caption{Performance under more rigorous attacks. Higher $\textcolor{blue}{\uparrow}$ and lower $\textcolor{red}{\downarrow}$ values indicate better performance.}
\label{tab:attacks}
\footnotesize
\vspace{-2pt}

% line 1
\begin{subtable}{\columnwidth}
\centering
\caption{Performance under adaptive attacks.}
\label{tab:whitebox}
\resizebox{\columnwidth}{!}{
\begin{tabular}{@{}l|c|c|c@{}}
\toprule[1.5pt]
Type & Unsafe Ratio (\%) \color{red}{$\downarrow$} & CLIP Score \color{blue}{$\uparrow$} & LPIPS Score \color{red}{$\downarrow$} \\
\midrule[1.5pt]
Vanilla & \textbf{6.25} & 26.10 & 0.697 \\
$\ell_\infty$-PGD on Embeddings Training & 9.13 & \textbf{26.16} & \textbf{0.696} \\
$\ell_\infty$-PGD on Gated Network Prediction  & 11.00  & 24.59 & 0.717 \\
\bottomrule[1.5pt]
\end{tabular}
}
\vspace{-2pt}
\end{subtable}

\vspace{2pt}

% line 2
\begin{minipage}{0.62\columnwidth}
\begin{subtable}{\textwidth}
\centering
\caption{Gated network behavior analysis.}
\label{tab:gated_analysis}
\resizebox{\textwidth}{!}{
\begin{tabular}{@{}c|cc|cc@{}}
\toprule[1.5pt]
\multirow{1}{*}{Metric} & \multicolumn{2}{c}{Acc. (\%) \color{blue}{$\uparrow$}} & \multicolumn{2}{|c}{mTS} \\
\midrule
\multirow{1}{*}{Dataset} & Malicious & Benign & Malicious \color{blue}{$\uparrow$} & Benign \color{red}{$\downarrow$} \\
\midrule[1.5pt]
Vanilla & \textbf{99.13} & \textbf{93.00} & \textbf{0.99} & \textbf{0.07} \\
$\ell_\infty$-PGD   & 69.63 & 81.50 & 0.59 & 0.34 \\
\bottomrule[1.5pt]
\end{tabular}
}
\end{subtable}
\end{minipage}
\hfill
\begin{minipage}{0.37\columnwidth}
\begin{subtable}{\textwidth}
\centering
\caption{Jailbreak attacks.}
\label{tab:jailbreak}
\resizebox{\textwidth}{!}{
\begin{tabular}{@{}c|c|c@{}}
\toprule[1.5pt]
Metric & \multicolumn{2}{c}{Unsafe Ratio (\%) \color{red}{$\downarrow$}} \\
\midrule
Dataset & CogMorph & RT-Attack \\
\midrule[1pt]
SDv1.4 & 13.75 & 8.40 \\
Ours & \textbf{1.25} & \textbf{0.42} \\
\bottomrule[1.5pt]
\end{tabular}
}
\end{subtable}
\end{minipage}

\vspace{-10pt}
\end{table}

\subsection{Defense on More Rigorous Attacks}

This part evaluates our defense on more rigorous attacks, including specially designed adaptive attacks and jailbreaks.

\textbf{Adaptive attack on defensive embeddings training.} We first implement targeted white-box adversarial attacks where attackers know details of our defense and can directly manipulate/break its process~\cite{liu2023towards,guo2023towards,liu2023x,liu2021training}. Specifically, we design an adaptive attack on defensive embedding training via $\ell_\infty$-bounded PGD~\cite{madry2019deep} ($\epsilon=0.3$, $\alpha=0.1$, $\text{iters}=20$), which perturbs the predicted noise residual under MSE loss to misalign the denoising trajectory, aligning vanilla model training setups (\eg, 500 training steps with $\lambda=0.7$). As shown in \Tref{tab:whitebox}: \ding{182} The unsafe ratio increases from 6.25\% to 9.13\%, revealing a clear weakening of defense capacity under $\ell_\infty$-PGD adversarial attack. \ding{183} LPIPS remains stable and CLIP slightly increases, indicating that benign preservation capability is retained despite adversarial attack. A possible reason lies in the gating mechanism, which dynamically modulates the strength of defense embeddings via interpolation. When not compromised by adversarial perturbation, this mechanism reliably ensures that safe prompts preserve high-quality benign generation.

\textbf{Adaptive attack on gated network prediction.} We further examine a stronger adaptive attack targeting the gated network, where $\ell_\infty$-bounded PGD perturbations are applied to the pooled text embedding under cross-entropy loss to mislead the toxicity classifier~\cite{liu2022harnessing,tang2021robustart,zhang2021interpreting,liu2020bias}. As shown in \Tref{tab:whitebox} and \Tref{tab:gated_analysis}, this results in substantial performance degradation across all indicators: unsafe ratio rises sharply from 6.25\% to 11.00\%, LPIPS worsens from 0.696 to 0.717, and CLIP drops from 26.10 to 24.59. These results highlight that disrupting the gated network significantly weakens both suppression capacity and benign preservation.
To further understand this degradation, we analyze the gated network classification and toxicity prediction behaviors under attack by measuring accuracy (Acc.) and mean predicted toxicity score (mTS) on malicious and benign datasets, as shown in \Tref{tab:gated_analysis}. \ding{182} After PGD attack, malicious accuracy drops sharply from 99.13\% to 69.63\%, and benign accuracy from 93.00\% to 81.50\%, indicating a substantial decline in the network ability to reliably distinguish between unsafe and benign inputs.
\ding{183} The mean toxicity score on malicious inputs drops from 0.99 to 0.59, while benign mTS increases from 0.07 to 0.34. Since high toxicity scores are desirable for malicious prompts (to trigger strong defense) and low scores for benign prompts (to retain quality), this reversed trend confirms that the gating signal becomes corrupted under attack. These findings demonstrate the pivotal role of the gated network in the overall defense framework.

\begin{figure} [!t]
    \centering
    \setlength{\abovecaptionskip}{1pt}% 
    \setlength{\belowcaptionskip}{1pt}%
    \includegraphics[width=1\linewidth]{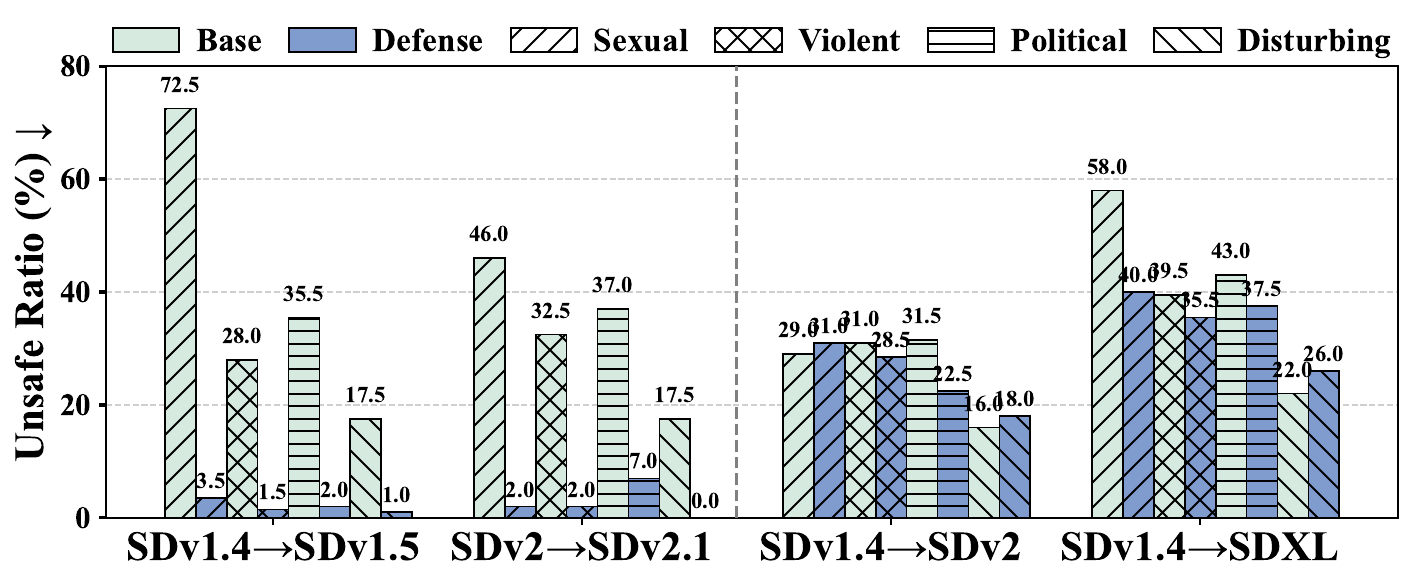}
    \caption{Transferability of \tool across T2I models with same or different text encoder architectures.}
    \label{fig:transfer_bar}
    \vspace{-13pt} % 
\end{figure}

\textbf{Jailbreak attack.} For jailbreak evaluation, we generate 200 harmful prompts via two representative jailbreak techniques (\eg, CogMorph~\cite{jing2025cogmorph} and RT-Attack~\cite{gao2024rtattack}). As shown in ~\Tref{tab:jailbreak}, compared to SDv1.4, our method consistently lowers the unsafe ratio across both attack types, with the largest drop reaching 12.5\% under CogMorph. This demonstrates that \tool exhibits strong robustness against prompt-level jailbreak attacks, benefiting from its training paradigm that leverages semantically separated textual pairs, enhancing resistance to adversarial manipulations in the input text.

\subsection{Ablation Studies}
In this part, we conduct ablation studies on some key components and parameters of our method.

\textbf{Loss balancing factor $\lambda$.}
We investigate the impact of the loss weight $\lambda$ that controls the balance between detoxification strength and benign preservation. \Tref{tab:lambda_ablation} shows the performance of models trained with different $\lambda$ values.
\ding{182} As $\lambda$ increases, the unsafe ratio drops significantly, from 66.5\% at $\lambda=0.1$ to 1.0\% at $\lambda=0.8$ and above, indicating stronger detoxification effect with heavier emphasis on safety alignment.
\ding{183} However, this comes at the cost of prompt-image fidelity. CLIP scores drop steadily with larger $\lambda$, and LPIPS increases, especially when $\lambda \geq 0.7$, reflecting semantic drift and visual degradation.
\ding{184} Overall, we find that $\lambda \in [0.6,0.7]$ provides the best trade-off, achieving strong suppression while maintaining generation quality.

\textbf{Inference strategies.} We evaluate four inference time strategies using the same trained model on Sexual testset, varying the number of soft tokens (\ie, Single or Combine) and whether gated network is applied (denoted as Dynamic), as shown in \Fref{fig:ablation_dynamic_combine}.
\ding{182} Single token strategies, particularly Single+Dynamic, improve CLIP fidelity (up to 26.46) but significantly weaken safety, indicating a trade-off between fidelity and suppression.
\ding{183} The default strategy (Combine+Dynamic) achieves the lowest unsafe ratio (1.99\%) and strong semantic fidelity (CLIP 26.34), offering the best safety–utility trade-off.
\ding{184} Removing the gated control leads to a higher unsafe ratio and a notable drop in CLIP score (22.55), confirming the importance of the gating mechanism.
These results suggest that Combine+Dynamic delivers the best trade-off between safety and fidelity, making it the recommended default. Additionally, other combinations can be selected flexibly to accommodate the diverse and nuanced safety requirements of real-world prompts.

\begin{table}[!t]
\centering
\renewcommand{\arraystretch}{1} % 
\setlength{\abovecaptionskip}{2pt}% 
\setlength{\belowcaptionskip}{2pt}%
\caption{Performance of \tool on Sexual category across different $\lambda$ at the setting of 500 training steps. Higher $\textcolor{blue}{\uparrow}$ and lower $\textcolor{red}{\downarrow}$ values indicate better performance.}
\label{tab:lambda_ablation}
\footnotesize
\resizebox{\columnwidth}{!}{ %
\begin{tabular}{@{}c|ccccccccc@{}}
\toprule[1.5pt]
Metric & \textbf{0.1} & \textbf{0.2} & \textbf{0.3} & \textbf{0.4} & \textbf{0.5} & \textbf{0.6} & \textbf{0.7} & \textbf{0.8} & \textbf{0.9} \\
\midrule[1.5pt]
Unsafe Ratio (\%)$\textcolor{red}{\downarrow}$ & 66.5 & 15.0 & 12.5 & 8.0 & 10.0 & 7.0 & 6.0 & 1.0 & 1.0 \\
\midrule
CLIP Score$\textcolor{blue}{\uparrow}$         & 26.33 & 25.96 & 25.20 & 24.54 & 24.45 & 24.37 & 24.23 & 20.34 & 21.17 \\
\midrule
LPIPS Score$\textcolor{red}{\downarrow}$        & 0.699 & 0.703 & 0.713 & 0.718 & 0.710 & 0.715 & 0.721 & 0.714 & 0.731 \\
\bottomrule[1.5pt]
\end{tabular}
}
\vspace{-2pt} % 
\end{table}

\begin{figure} [!t]
    \centering
    \setlength{\abovecaptionskip}{1pt}% 
    \setlength{\belowcaptionskip}{1pt}%
    \includegraphics[width=1\linewidth]{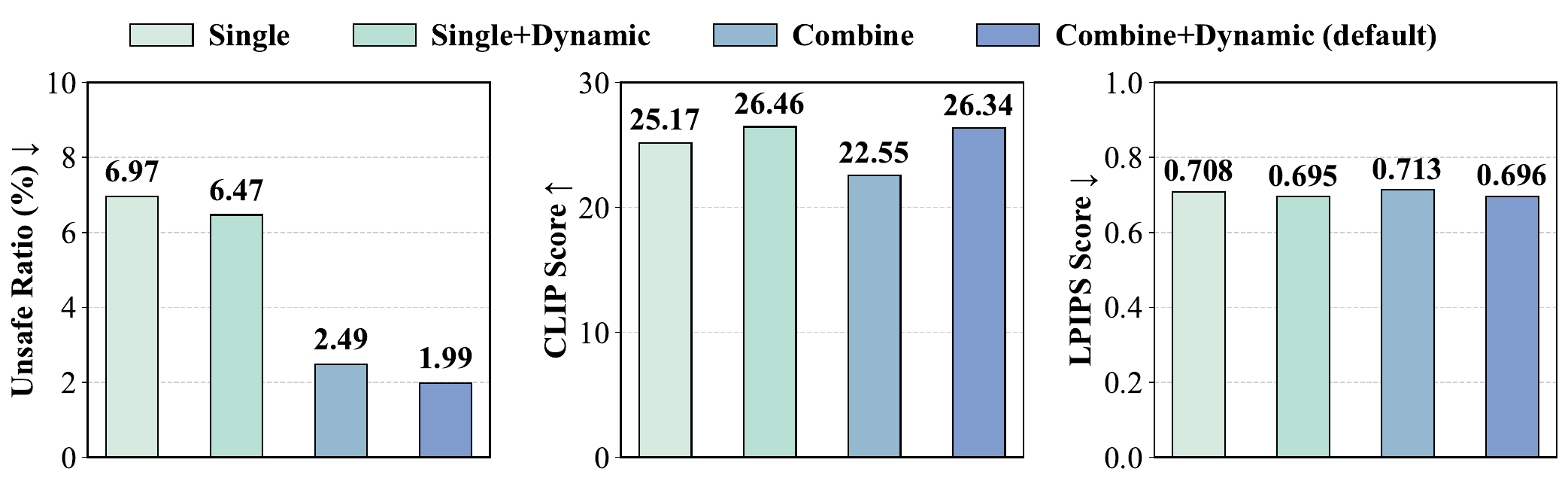}
    \caption{Ablation on different inference strategies.}
    \label{fig:ablation_dynamic_combine}
    \vspace{-10pt} % 
\end{figure}

\section{Conclusion and Future Work}
This paper introduces \toolns, a lightweight and effective soft prompt–based defense framework for mitigating NSFW content generation in T2I models. Extensive experiments show that \tool achieves state-of-the-art suppression across multiple NSFW categories, while maintaining high fidelity for benign content and generalizing well across models and domains. \textbf{Future work.} \ding{182} We plan to refine adaptive control with more fine-grained toxicity estimation. \ding{183} We also aim to extend our framework to tasks like image inpainting and text-to-video generation.
{
    \small
    \bibliographystyle{ieeenat_fullname}
    \bibliography{main}
}


\end{document}